\documentclass[letterpaper, 10 pt, journal, twoside]{IEEEtran}

\IEEEoverridecommandlockouts                              

\usepackage{graphicx} 
\usepackage{amsmath} 
\usepackage{amssymb}  
\usepackage{bm}
\usepackage{subfig}
\usepackage{booktabs}
\usepackage[hyphens]{url}
\usepackage[hidelinks]{hyperref}
\usepackage{cleveref}
\usepackage{siunitx}
\usepackage[mode=buildnew]{standalone}
\usepackage{tikz}
\usepackage{pgfplots}

\usepackage{flushend}

\usepackage[symbol]{footmisc}

\usepackage[colorinlistoftodos,prependcaption,textsize=small]{todonotes}

\usepackage{color}

\newcommand{\aes}[1]{\textcolor{black}{#1}}
\newcommand{\md}[1]{\textcolor{black}{#1}}
\newcommand{\mm}[1]{\textcolor{black}{#1}}

\newcommand{\aesf}[1]{\textcolor{black}{#1}}
\newcommand{\mdf}[1]{\textcolor{black}{#1}}
\newcommand{\mmf}[1]{\textcolor{black}{#1}}
\newcommand{\rkkf}[1]{\textcolor{black}{#1}}

\renewcommand{\baselinestretch}{0.97}
\usepackage[font=small,skip=3pt]{caption}

\begin{document}


\title{\rkkf{Sim-to-Real for Soft Robots using Differentiable FEM:\,Recipes\,for\,Meshing,\,Damping,\,and\,Actuation}}

\author{Mathieu Dubied$^{1}$, Mike Y. Michelis$^{1}$, Andrew Spielberg$^{2}$, and Robert K. Katzschmann$^{1}$%
\thanks{Manuscript received: September 13, 2021; Revised December 14, 2021; Accepted January 27, 2022.}
\thanks{This paper was recommended for publication by Editor Cecilia Laschi upon evaluation of the Associate Editor and Reviewers' comments.
} 
\thanks{$^{1}$Mathieu Dubied, Mike Y. Michelis, and Robert K. Katzschmann are with the Soft Robotics Lab, Department of Mechanical and Process Engineering, ETH Zurich, Switzerland
        {\tt\footnotesize \{\href{mailto:mdubied@ethz.ch}{mdubied}, \href{mailto:mmichelis@ethz.ch}{mmichelis}, \href{mailto:rkk@ethz.ch}{rkk}\}@ethz.ch}}%
\thanks{$^{2} $Andrew Spielberg is with the Computer Science and Artificial Intelligence Laboratory, MIT, Cambridge, MA, USA
        {\tt\footnotesize \href{mailto:aespielberg@csail.mit.edu}{aespielberg@csail.mit.edu}}}%
\thanks{Digital Object Identifier (DOI): see top of this page.}
}

\markboth{IEEE Robotics and Automation Letters. Preprint Version. Accepted January, 2022}
{Dubied \MakeLowercase{\textit{et al.}}: Sim-to-Real for Soft Robots using Differentiable FEM: Recipes for Meshing, Damping, and Actuation} 


\maketitle


\begin{abstract}

An accurate, physically-based, \rkkf{and differentiable}
  model of soft robots can unlock downstream applications in optimal control.  
The Finite Element Method (FEM) \md{is an} expressive approach for modeling highly deformable structures such as dynamic, elastomeric soft robots. 
In this paper, we compare virtual robot models simulated using \rkkf{differentiable FEM} with measurements from their physical counterparts. In particular, we examine several soft structures with different morphologies: a clamped \rkkf{soft} beam under external force, a pneumatically actuated soft robotic arm, and a soft robotic fish tail. 
We benchmark and analyze different meshing resolutions and elements (tetrahedra and hexahedra), numerical damping, and the \aesf{efficacy of differentiability for parameter calibration} \aes{\aesf{using a simulator based on the fast Differentiable Projective Dynamics} (DiffPD)}.  
We also advance \mdf{FEM modeling} \aes{in application to soft robotics} by proposing a predictive model for pneumatic soft robotic actuation.  
Through our \rkkf{recipes and }case studies, we provide strategies and algorithms for matching real-world physics in simulation, making \mdf{FEM} useful for \md{soft} robots.
\footnote[2]{Please see \url{https://github.com/srl-ethz/diffPD_sim2real} to download and run the virtual experiments.}


\end{abstract}


\begin{IEEEkeywords}
Modeling, Control, and Learning for Soft Robots, Dynamics, Optimization and Optimal Control, Simulation and Animation
\end{IEEEkeywords}

\section{Introduction}

\IEEEPARstart{S}{oft} robotics promises the rise of a new generation of robots with built-in compliance, damping, elastic energy storage, continuous actuation, and other morphologically-encoded behavioral features. 
 However, one of the key challenges in realizing the potential of this field is to develop a simulation model that is useful for analyzing these effects and developing real-world controllers. 
 \aesf{The Finite Element Method (FEM), when combined with highly stable implicit Euler integration,} shows promise for real-world soft robotic models due to its speed and accuracy.  Furthermore, \aesf{the} differentiability \aesf{of some FEM models} \cite{Du2021DiffPD:Dynamics} enables the fast and accurate computation of gradients for efficient optimization of a robot's design and control.
 
    

\aes{\aesf{This approach, however}, introduces several challenges when applied to soft robotic modeling.}  \md{Although \aesf{the implicit Euler integration} is unconditionally stable, it also causes significant numerical damping} --- a side effect that causes simulated structures to exhibit mechanical damping, even when mechanical damping is not an explicit feature of the material model.  Moreover, despite \aesf{FEM} being based on continuum mechanics, it attempts to approximate continuous phenomena \emph{via} discretization.  The choice of this discretization can have serious ramifications on the model's accuracy. These challenges must be addressed \aes{for \aesf{FEM} to be useful to the robotics community.}

\begin{figure}
    \centering
    \includegraphics[width=1\linewidth]{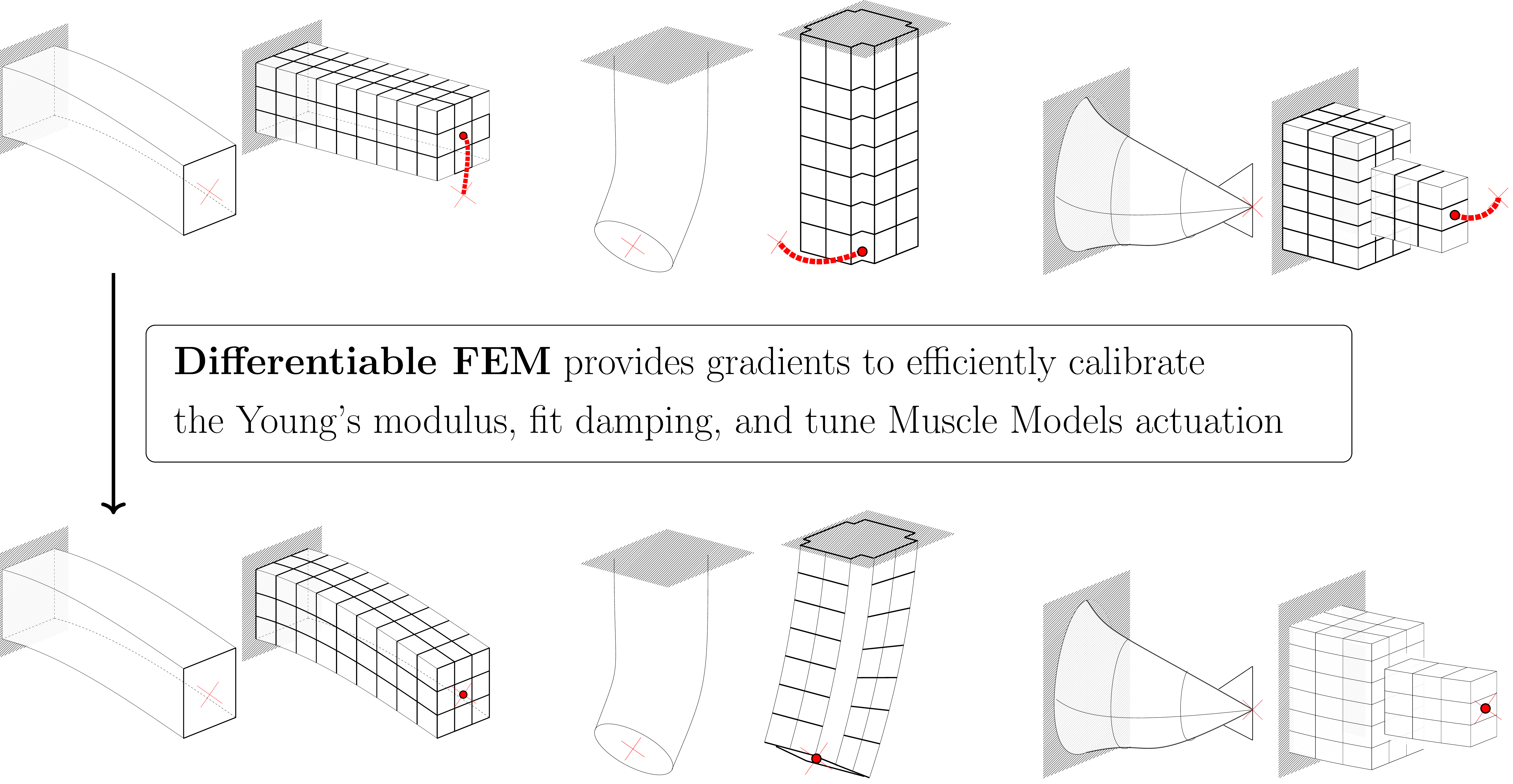}
    \caption{\mdf{We show that differentiable FEM can be used to improve models and decrease modeling error, bringing simulation closer to reality.}}
    \label{fig:summary}
\end{figure}
 
 

In this paper, we evaluate the ability of \aesf{dynamic FEM} to model 
deformable soft robots with \rkkf{fluidic} actuation.  \aesf{In particular, we employ Projective Dynamics (PD), an FEM formulation that can be made efficiently differentiable}.  \aes{We rely on the differentiable nature of the Differentiable Projective Dynamics (DiffPD) framework \cite{Du2021DiffPD:Dynamics} to overcome the sim-to-real gap}.  Our three contributions toward this goal are:

 \begin{enumerate}
    \item \emph{Benchmarking \aesf{FEM (in particular,} PD) against reality for different meshes}.  We perform characterizing experiments on three different soft robotic structures, and compare the measurement data to the PD simulations.  \md{We consider different mesh resolutions and two different mesh element types (tetrahedra and hexahedra).  We also benchmark the obtained results against a commercial FEM solution.}
    
    \item \emph{Characterizing \rkkf{and adapting} numerical damping \rkkf{to match} material damping \rkkf{in reality}}. We characterize the numerical damping, which originates from the implicit Euler scheme. We benchmark this numerical damping against an analytical solution we derive. We then propose a new method to adjust the numerical damping so that it matches the material damping.  
    \item \emph{Creating FEM models \rkkf{with} soft robotic actuation \rkkf{matching reality}}. We compare PD simulations to reality for geometrically complex \mm{soft robotic} actuators. Since a precise model of the pressure chambers leads to unsatisfactory results in terms of speed and accuracy, we present ``Muscle Models'' as an \md{alternative} 
    means of modeling \rkkf{fluidic} chambers.
\end{enumerate}

These contributions address the most challenging aspects \rkkf{when} simulating robots \aes{using \aesf{FEM}} \rkkf{and transferring them to reality}.
Thus, we provide a guide for practitioners \rkkf{to} quickly simulate and optimize real-world soft robots in the future.


\section{Related Work}\label{related_work}

In this section, we begin by describing existing methods for simulating soft robots. We then summarize related progress toward overcoming their sim-to-real gap.

\subsection{Soft Robotic Simulation}
Many approaches for accurately simulating soft robots have been proposed over the years.
One approach is to use a model based on simplifying assumptions, such as the Augmented Rigid Body model~\cite{della2018dynamic} that is based on the Piecewise Constant Curvature model~\cite{RobertJ.Webster2010DesignReview:}. A second approach is to develop data-driven models that are based on measurements that have been performed on the \rkkf{robot}~\cite{Haggerty2020, Bruder2019, han2021desko}. A third approach is to employ models that discretize the robot's continuum mechanics and solve the equations of motions for every element of the discretization~\cite{Tonkens2020}. 

In this work, we focus on discretization approaches, specifically FEM. In the field of soft robotics, the \aes{Simulation Open Framework Architecture (SOFA)} is widely used. SOFA accurately models highly deformable structures, and it can be used as a basis for fast simulations that use both traditional and reduced order simulation methods \cite{ Tonkens2020, Duriez2017FrameworkModel, Goury2018FastReduction}.  Other promising simulation frameworks have recently emerged, such as the DiffPD framework \cite{Du2021DiffPD:Dynamics}, which is based on PD \cite{Bouaziz2014}. 
PD employs a projection-based implicit Euler method to solve the equations of motion and to achieve more stable and efficient simulations than explicit schemes (although this comes at the expense of accuracy). It has been optimized for performance over recent years \cite{Wang2015ADynamics, Wang2016DescentGPU, Fratarcangeli2016Vivace:Dynamics}. 
DiffPD, and other simulators \cite{Chen2017Dynamics-awareDesign,Hu2019ChainQueen:Robotics}, can even be designed to handle contact both accurately and efficiently.

Notably, DiffPD extends PD to the family of simulators that are \emph{differentiable}. Differentiable simulators allow for model-based computation of gradients of any variable (state, model, or control) in a system with respect to any other variable. Differentiable simulators have been
used to efficiently solve control and design optimization tasks in soft-body regimes 
 \cite{Hu2019ChainQueen:Robotics, Hahn2019Real2Sim:Motion, Hu2019DiffTaichi:Simulation,Geilinger2020ADD:Contact}.  %

\subsection{Sim-to-Real for Soft Robotics}
As the field of soft robotics has grown, so too has interest in translating simulated results to the physical world.  In the case of linear manipulators such as arms, simplified models have been effective.  For example, soft robots modeled as rigid links with torsional spring joints have been used to motion plan soft robotic arms through real-world maze-like environments \cite{marchese2016design, marchese2016dynamics}.  More complex PCC models have also been used to create dynamic soft robotic arms with optimal control \cite{della2018dynamic}, which translate faithfully to reality.

Finite element models have long garnered special attention, however, due to their ability to scale to more complex geometry.
COMSOL, Abaqus, and other commercial and open-source FEM packages have been used for the modeling of soft structures and robots. 
These programs have been particularly useful for designing physical soft robotic structures \cite{mosadegh2014pneumatic, joyee2019fully}.
SOFA \cite{allard2007sofa} have successfully been deployed in a number of applications of kinematic control of real-world soft robots, including real-time control \cite{duriez2013control, zhang2016kinematic, coevoet2017optimization}. Similar work has 
demonstrated dynamic analysis and control of soft robots by relying on reduced order models for real-time tractability \cite{Goury2018FastReduction, Katzschmann2019DynamicallyObserver}. Other relevant work includes demonstrations of trajectory optimization and control on physical soft walking robots and grippers \cite{bern2017fabrication, bern2019trajectory}. While the above listed work takes steps to faithfully model soft robots and compare physical and virtual performance, they do not investigate systematic approaches and analyses needed in order to put the virtual and physical models into correspondence; this step is especially important but also difficult when dealing with dynamic robots, and it is often not treated in-depth.  

The research works most similar to ours (namely, the works that take systematic approaches to overcoming the sim-to-real gap) include \cite{kriegman2020scalable} and \cite{du2021underwater}.
A data-driven approach to building complex voxel-based soft robots\rkkf{~\cite{kriegman2020scalable}} tunes parameters for simpler designs and scales up complexity using a spring-based simulator \cite{hiller2014dynamic}. Repeatedly applying system identification \rkkf{of} model \emph{parameters} in the DiffPD framework \rkkf{improves} the performance of robotic swimmers, but does not address issues inherent to the model itself\rkkf{~\cite{du2021underwater}}. 
Our work analyzes the important nuances of soft robotic modeling, specifically in a PD framework, examining the effects of design choices such as meshing, \md{material} and numerical damping, and actuation models.

\section{Theoretical Background}\label{theoretical_background}
\subsection{Soft Robotic Finite Element Model}\label{finite_element_method}

\md{PD, like} \rkkf{other} \md{classical FEM solvers,} solves the PDEs specified by continuum mechanics through spatial and time discretization.

\subsubsection{Spatial Disc\md{r}etization}\label{spatial_discretization}
Starting from a continuous body with an infinite number of DOFs, \md{the spatial discretization in finite elements (FE)} reduces the system's dimensionality to a tractable, finite number. The shape of the body is approximated by a mesh of finite elements, each spanned by $n$ nodes. \md{For volumetric elements in 3D space, each node has 3 translational DOFs} and is described by its position and velocity vectors through Newton's second law:
\begin{equation}\label{EOM}
    \bm{M}\dot{\bm{v}}=\bm{f}_{\text{ext}}+\bm{f}_{\text{int}}(\bm{q})
\end{equation}

\noindent where $\bm{q}=\bm{q}(t)\in \mathbb{R}^{3n}$ describes the position of every node, $\bm{v}=\bm{v}(t)\in \mathbb{R}^{3n}$ describes the nodes' velocities, $  \bm{M}\in \mathbb{R}^{3n\times 3n}$ is the mass matrix, $\bm{f}_{\text{ext}}=\bm{f}_{\text{ext}}(t)\in \mathbb{R}^{3n}$ accounts for external forces, and  $\bm{f}_{\text{int}}(\bm{q})=\bm{f}_{\text{int}}(\bm{q},t)\in \mathbb{R}^{3n}$ accounts for internal forces. \mdf{We note that the formulation for PD's internal forces does not include system velocities; thus, velocity-dependent forces such as damping forces are modeled as \emph{external} forces (see \Cref{modeling_damping}).} 


\subsubsection{Time Discretization}\label{time_discretization}
PD and \md{classic} FEM solvers consider \eqref{EOM} for discrete time steps by, for example, using the implicit Euler method. The implicit Euler method approximates the derivative of a function $u_i=u(x_i, t_i)$ as
\begin{equation}\label{backward_Euler}
    \frac{\partial u}{\partial t}(x_{i}, t_{i}) = \dot{u}(x_{i}, t_{i}) \approx \frac{u_{i+1}-u_i}{h}
\end{equation}

\noindent where $h$ is the time step defined as $h=t_{i+1}-t_i$ and $i=0,1,2,...$. Applying the \md{implicit} Euler method to \eqref{EOM}, we obtain the following discretized equations of motion:
\begin{equation}\label{EOM_discretized}
    \begin{aligned}
        \bm{q}_{i+1} &= \bm{q}_i + h \bm{v}_{i+1} \\
        \bm{v}_{i+1} &= \bm{v}_i + h \bm{M}^{-1}(\bm{q})[\bm{f}_{\text{int}}(\bm{q}_{i+1})+\bm{f}_{\text{ext}}]
    \end{aligned}
\end{equation}

\subsection{Projective Dynamics}\label{diffpd}

\subsubsection{Optimization Formulation}\label{optimization_problem_formulation}
PD formulates \eqref{EOM_discretized} as an optimization problem to solve this equation efficiently:
\begin{equation}\label{optimization_problem}
    \min_{\bm{q}_{i+1}}\left\{ \frac{1}{2h^2}||\bm{M}^{\frac{1}{2}}(\bm{q}_{i+1}-\bm{y}_i)||_2^2 +E_{\text{int}}(\bm{q}_{i+1})\right\}
\end{equation}

\noindent with $\bm{f}_{\text{int}}(\bm{q}_{i+1})= - \nabla E_{\text{int}}(\bm{q}_{i+1})$ and $\bm{y}_i=\bm{q}_i+ h \bm{v}_i + h^2\bm{M}^{-1}\bm{f_{\text{ext}}}$. 
The first term is the momentum potential energy and the second term $E_{\text{int}}$ is the elastic potential energy.


\subsubsection{Elastic Energy Formulation}\label{elastic_energy_formulation}
In nonlinear continuum mechanics, the elastic energy $E_{\text{int}}(\bm{q}_{i+1})$ is a nonlinear function \aes{that serves as a restorative spring-like potential energy that attracts each state to its undeformed state. Such potential energies can yield linear or nonlinear spring-like forces.  Some common potential energies for modeling are described in \cite{Sifakis2012FEMReduction}.}
PD introduces an elastic energy in a form that is convenient for efficient computation. \aes{The elastic energy's form is introduced in \cite{Bouaziz2014ProjectiveSimulation} and is also used by DiffPD \cite{Du2021DiffPD:Dynamics}.}

\subsection{Material Damping}\label{material_damping}
Damping is a phenomenon that reduces the internal velocity of a system by dissipating energy. It is commonly associated with amplitude oscillatory responses.

With \textit{material} damping, we refer to the damping scenario in which the source of damping is physical, and not numerical. Numerical damping is discussed in \Cref{numerical_damping}.

\subsubsection{Single DOF System and Damping Ratio}
The prototypical equation of motion of a single DOF system undergoing free vibration can be expressed as
\begin{equation}\label{single_DOF}
    \ddot{x}+2\zeta\omega_0\dot{x}+\omega_0^2 x=0
\end{equation}

\noindent with damping ratio $\zeta$ and undamped frequency $\omega_0$. 

We can classify the solutions $x=x(t)$ of \eqref{single_DOF} depending on the value of the damping ratio $\zeta$. Important for this paper are underdamped vibrations, where $0\leq\zeta<1$. 

\subsubsection{Underdamped Vibrations}\label{underdamped_vibrations}
In the case of underdamped vibrations, the solution of \eqref{single_DOF} can be written as

\begin{equation}\label{underdamped_sol}
    x(t) = e^{-\zeta\omega_0 t}\left[C_1 \cos(\omega_d t) + C_2\sin(\omega_d t)\right]
\end{equation}

\noindent where $\omega_d=\omega_0\sqrt{1-\zeta^2}<\omega_0$ is the damped natural frequency.

As an inverse problem, it is also possible to determine $\zeta$ based on data. We first need to compute the logarithmic decrement $\delta$, which is defined as follows:

\begin{equation}\label{log_dec}
    \delta=\frac{1}{m}\cdot \ln\left(\frac{X_k}{X_{k+m}}\right)
\end{equation}

\noindent \md{where $X_k$ is the $k$-th peak amplitude, and $X_{k+m}$ is the amplitude of the peak $m$ periods afterward. In this paper, we consider $m\in\{1,2,3,4\}$ depending on the time step $h$ of our simulations.}
From the logarithmic decrement $\delta$, one can determine the damping ratio $\zeta$:

\begin{equation}\label{damping_ratio}
    \zeta=\frac{\delta}{\left(4\pi^2+\delta^2\right)^{1/2}}
\end{equation}

\subsection{\mdf{Numerical 
Damping}}\label{numerical_damping}
Unlike material damping which occurs due to physical phenomena (such as viscous drag, heat exchange, etc.), numerical damping occurs because of the time discretization method we use. The implicit Euler method, used in PD, is known to be \mdf{A-}stable but it also induces numerical damping \cite{Ammari2018NumericalEquations, Chen2018ExponentialSimulation}.

We illustrate this \mdf{fact}
through a simple example. 
 \mdf{We first define the following linear system:}

\mdf{
\begin{equation}\label{numerical_damping_example}
    \begin{cases}
      \frac{dx}{dt} = i\omega x(t),\quad \omega\in \mathbb{R}, \;t\in [0, +\infty)\\
      x(0)= x_0=1
     \end{cases}
\end{equation}}

The exact solution of this equation is an oscillation whose amplitude does not change over time:
\begin{equation}\label{numerical damping_example_exact_solution}
    x(t)=e^{i\omega t}=\cos(\omega t) + i\sin(\omega t)
\end{equation}

\mdf{The solution obtained using the implicit Euler method is}

\mdf{
\begin{equation}
    \frac{x_{j+1}-x_j}{h}=i\omega x_{j+1} \;\Rightarrow \; x_{j+1}=\frac{1}{1-hi\omega}x_j, \;\text{for } j\in \mathbb{N}
\end{equation}}

\mdf{Using $x_0=1$, we can rewrite this relation as}

\mdf{
\begin{equation}\label{euler_induction_formula}
    x_j= \left(\frac{1}{1-hi\omega}\right)^j \cdot x_0 = \left(\frac{1}{1-hi\omega}\right)^j
\end{equation}}

We can observe that for a finite value of $h$ and $\omega\in \mathbb{R}^+$, the amplitude of the oscillation is decreasing over time:

\begin{equation}\label{decreasing_amplitude}
    |x_j|= \left|\left(\frac{1}{1-hi\omega}\right)^j\right| \rightarrow 0 \quad\text{for } j\rightarrow \infty
\end{equation}

This phenomenon is called \textit{numerical} damping, as the source of the damping is purely an artifact of the numerical method that is used for time discretization. 

\section{Modeling: Meshing, Damping, and Actuation}\label{modeling}

\subsection{Meshing}\label{meshing}
In this work, we investigate two FE types for meshing: hexahedra and tetrahedra. \aes{Hexahedral meshes are traditionally locked to axis-aligned voxel grids in their rest configuration.  This leads to aliasing when one attempts to fit them to curved or angled surfaces}. However, \aes{while more geometrically accurate}, meshes based on linear tetrahedra suffer from \emph{locking} when used to model (nearly) incompressible materials ($\nu\rightarrow 0.5$) \cite{Babuska1992LockingProblems} or when applied to systems with highly nonlinear dynamical behavior \cite{Puso2006ATetrahedral}. These meshes yield a much stiffer behavior of the simulated structure than the real equivalent. 


\subsection{Damping}\label{modeling_damping}
Numerical damping is inherently present when \aes{using the} implicit Euler \aes{method} \aes{ (as  in PD)} 
 and can be \aes{quite} large (cf. \Cref{results_beam}). \aes{This can lead to large error in the simulated dynamic motion. In order to compensate for this error and }
fit the dynamic behavior of real structures, we model velocity-dependent state forces that \aes{correct} for numerical damping:
\begin{equation}\label{damping_compensation_formula}
    f_i^l=\Lambda\cdot v_{i-1}^{l}
\end{equation}
\noindent  where $\Lambda\in \mathbb{R}$ is a tuning parameter and $l$ the vertex to which the force is applied. 
We use the simulated velocity $v_{i-1}^{l}$ of the previous time step to compute the current state force $f_i^l$. \mdf{We analyse in \Cref{results_beam} the effect of this force on the stability of the system. If the value of $\Lambda$ is correctly chosen, the simulation results remain stable.}  
\subsection{Actuation}\label{actuation}

We design ``Muscle Models'' in order to predict the behavior of elastic pneumatics. As will be shown, these are a helpful alternative to the exact modeling of pressure chambers. We employ a muscle energy \cite{Du2021DiffPD:Dynamics, min2019softcon} which adds an additional energy term  to $E_{\text{int}}$ in \eqref{optimization_problem}; this term can be understood as a spring-like energy that is attributed to the mesh's elements:
\begin{equation}
    E(\bm{q})=\frac{w}{2}||(1-a)\bm{F}\bm{m}||^2
\end{equation}

\noindent  where $w$ can be \md{interpreted} as the stiffness of the spring, $a\in \mathbb{R}$ is the actuation signal, $\bm{F}$ is the deformation gradient, and $\bm{m}$ is the direction of the actuation. If $a>1$, extension occurs; if $a<1$, contraction occurs. 
\aes{While springs-like models are not pneumatic models as derived from first principles, they are much simpler to model in a PD simulator, and, as we demonstrate in section \Cref{results_arm} can be tuned to provide dynamic behavior that closely matches reality.}

\section{Experimental and Simulation Setups}\label{experimental_simulation_setups}

\subsection{Experimental Setup}\label{experimental_setup}

In this section, we discuss the three deformable structures that will serve as a testbed for our analysis.

\subsubsection{Clamped Beam under External Force}\label{materials_and_methods_beam}
\begin{figure}[tb]%
    \centering
    \subfloat
    {{\includegraphics[width=0.4\linewidth]
    {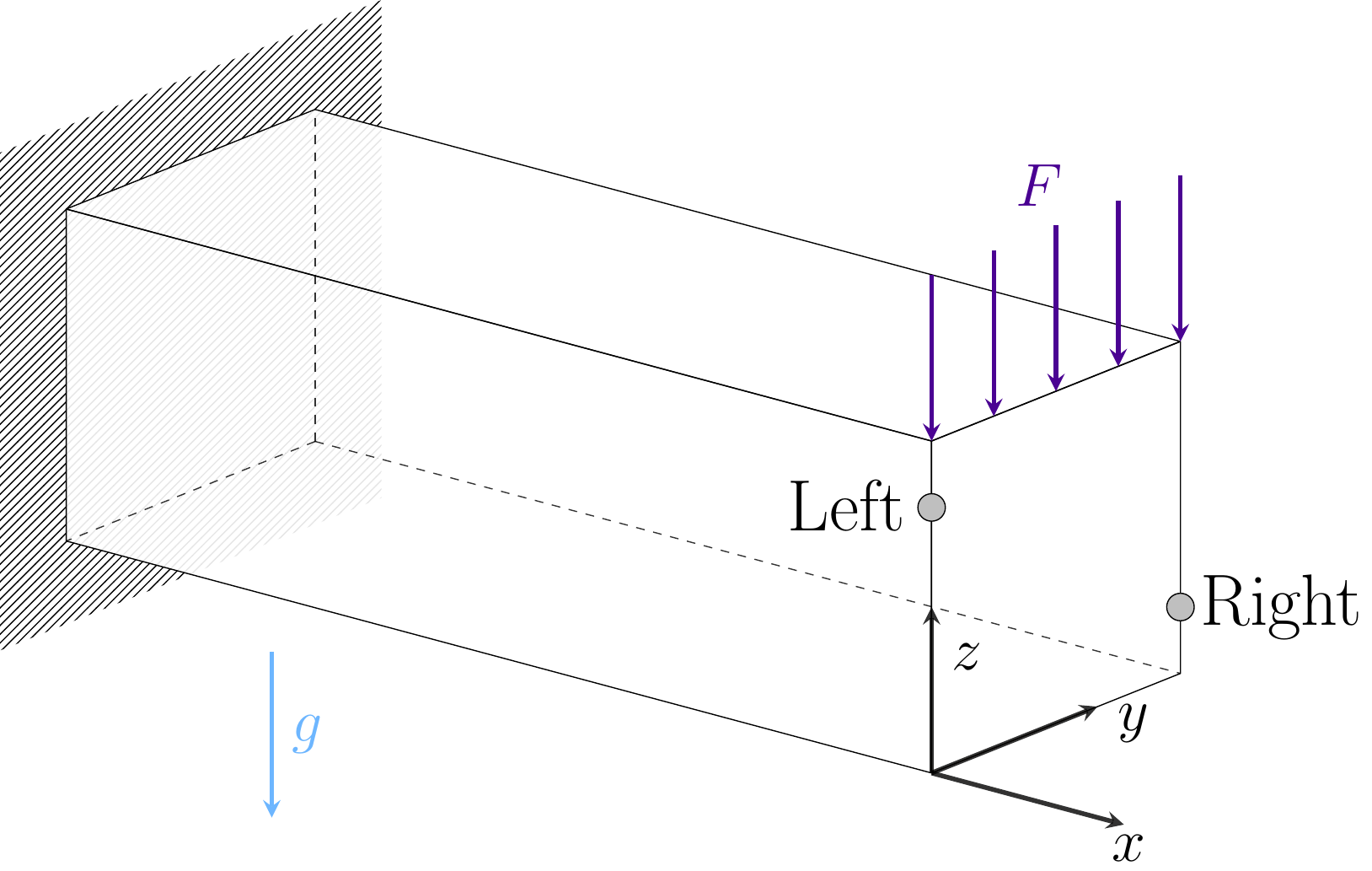}}}%
    \qquad
    \subfloat
    {{\includegraphics[width=0.4\linewidth]{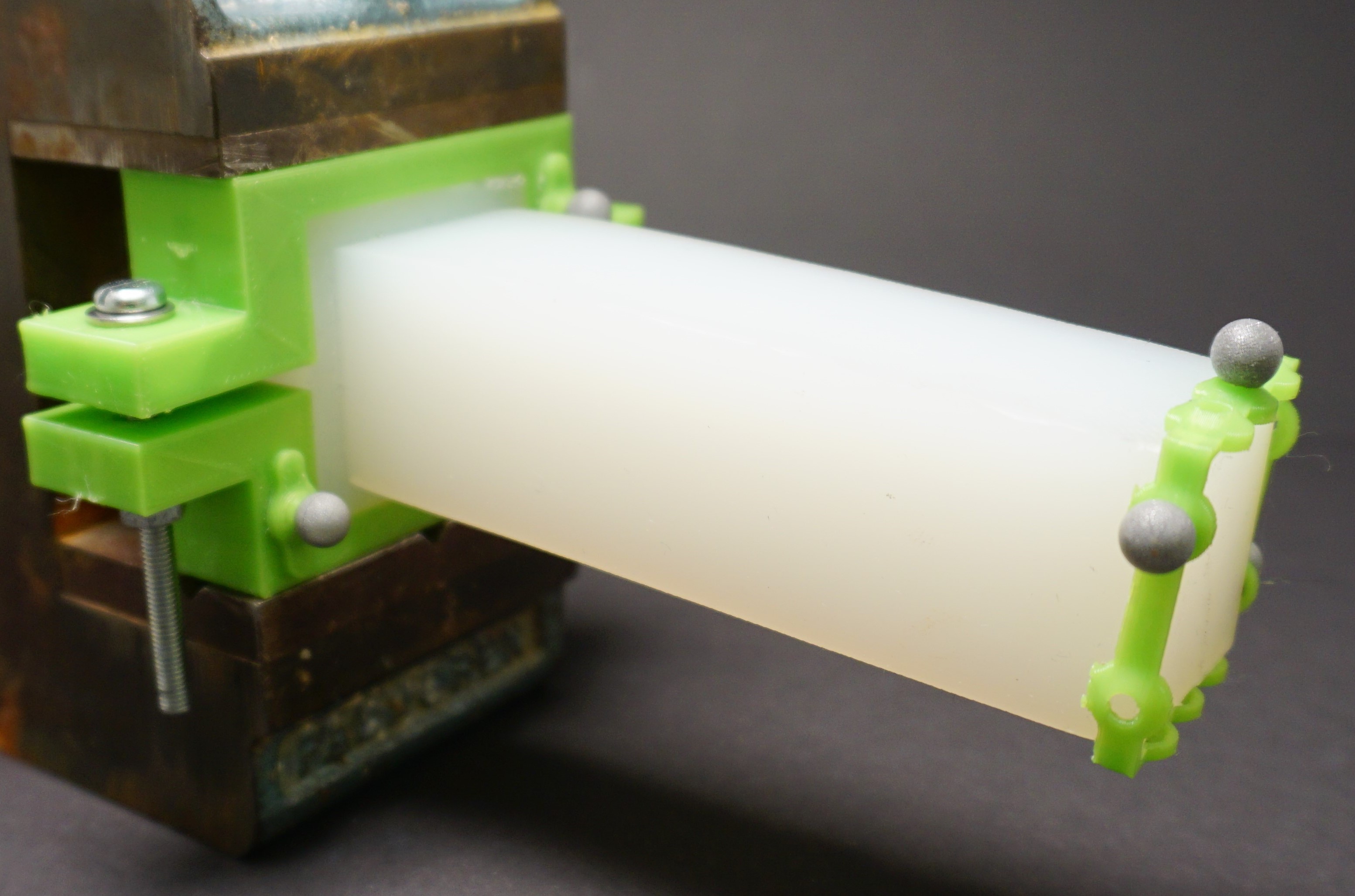}}}%
    \caption{Achematic with edge forces (left) and a photo (right) of the setup, including motion markers, for the clamped beam scenario.}%
    \label{fig:beam_scheme}%
\end{figure}

To benchmark the accuracy of PD, we begin by considering a simple, yet important structure: a clamped beam made of a highly deformable silicone elastomer  (Fig. \ref{fig:beam_scheme}).

We consider two scenarios: \emph{1)} we keep the beam at its horizontal starting position and then release it so it bends downwards under gravity; and \emph{2)} in addition to gravity, we apply an external force $F$ along the edge of the tip. To track the two reference points (``Left'' and ``Right''), we use a Motion Capture System from Qualisys.  See Fig. \ref{fig:beam_scheme} for details.

\subsubsection{Soft Robotic Arm}\label{materials_and_methods_arm}
The second structure is a pneumatically actuated soft robotic arm made of a silicone elastomer.

\begin{figure}[!b]
    \centering
    \subfloat[\centering Soft Arm]{{\includegraphics[width=0.5\linewidth]{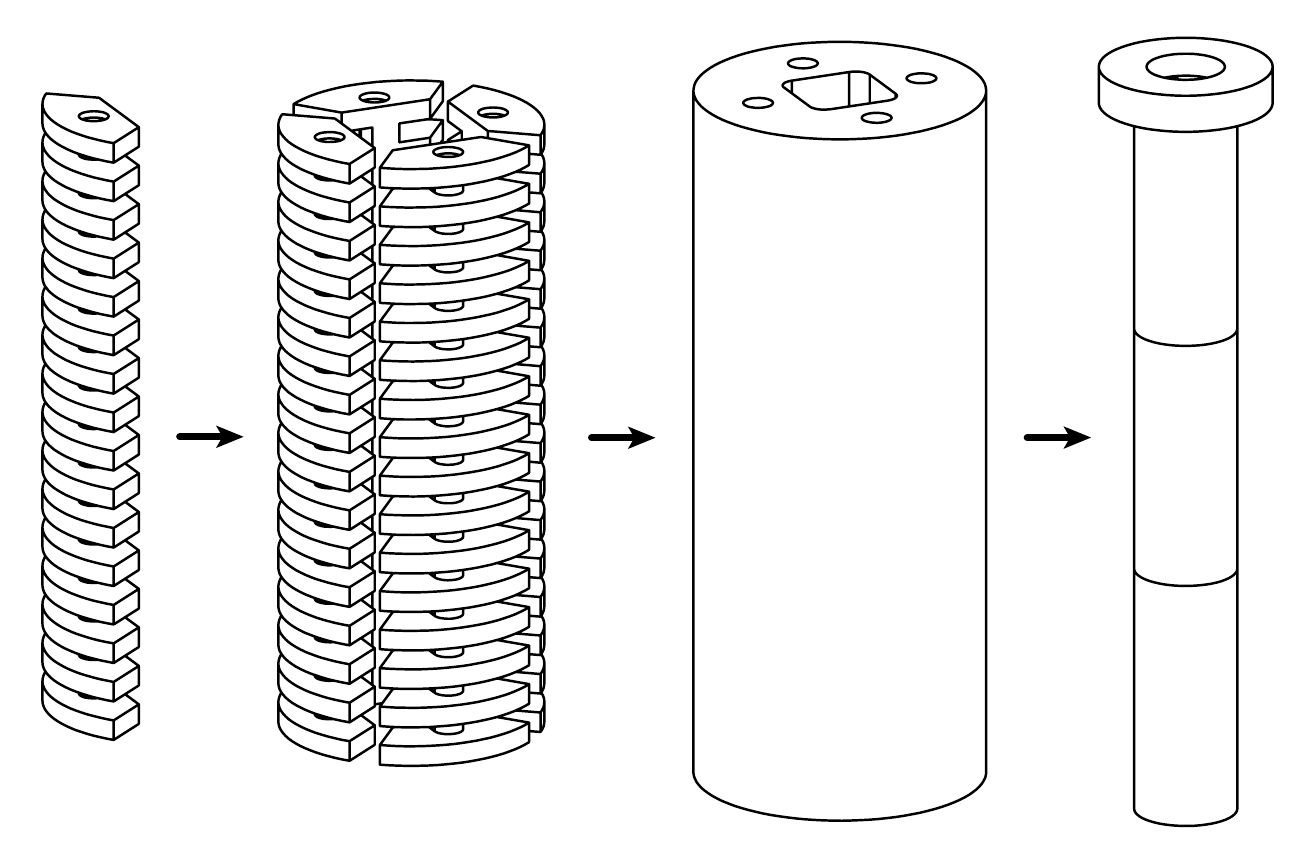}}}%
    \qquad
    \subfloat[\centering Soft Fish]{{\includegraphics[width=0.2\linewidth]{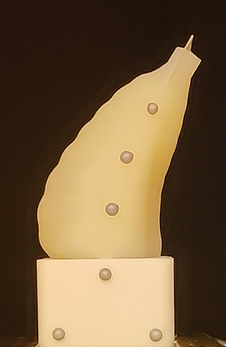}}}%
    \caption{(a) Arm segment with ribbed pressure chambers \cite{Katzschmann2019DynamicallyObserver}. (b) Soft fish tail.}
    \label{fig:arm_experimental_setup}
\end{figure}

Our arm is comprised of a single segment of the soft robotic arm described in \cite{Katzschmann2019DynamicallyObserver}. It has 4 pressure chambers, which have a complex geometry, as shown in Fig. \ref{fig:arm_experimental_setup}. We place motion markers on the robotic arm to track the tip's center position.

\subsubsection{Soft Fish Tail}\label{materials_and_methods_fish}
The last structure we consider is a pneumatic soft fish tail (Fig. \ref{fig:arm_experimental_setup}). We focus on the position of the last motion marker placed on the tail's spine.
\subsection{Simulation Setup}\label{simulation_setup}

In this section we explain our simulation setup.  For the remainder of this paper, unless specified otherwise, our simulations use a time step $h$ of $0.01$ \si{s}.

\subsubsection{Material Parameters}
The silicone elastomer used for the beam is Smooth-On Dragon Skin 10 with Shore Hardness 10A. For silicones, we assume incompressibility, \textit{i.e.}, $\nu=0.5$ ($\nu=0.499$ to avoid numerical singularities). The material density specified by Smooth-On is $\rho=1070$ \si{kg.m^{-3}}.



\md{Precise knowledge of the material properties is needed to achieve an accurate simulation. This is difficult with elastomers, which generally show large variability due to their fabrication process. We use COMSOL as a ground truth, but we will show in \Cref{results_beam} that DiffPD can also be used to calibrate material parameters using its differentiability.}

We simulate Case E (see \Cref{tab:beam_load_scenarios}) and tune the Young's modulus so that the static solution from COMSOL matches the real data. We choose Case E because the large load $F$ is dominant in the final deformation state, and the effects of setup inaccuracies are negligible. This method leads to 
a Young's modulus of $E=263824$ \si{Pa}. This value is similar to the values found in related literature \cite{Rothemund2018AActuators, Connolly2017AutomaticMatching}.

\subsubsection{Material Models }
 We simulated our structures, experimenting with \md{corotational linear elastic} and Neo-Hookean material models \cite{Sifakis2012FEMReduction}. \md{In our experiments, both models yield similar results; thus we only present results obtained with the corotational linear elastic model, which is numerically simpler to employ.}


\subsubsection{Optimization algorithm}\label{optimization_algorithm} \md{When optimizing any variable $\alpha$ in DiffPD, the following steps are taken. First, we choose an initial guess for $\alpha$ randomly. Then, we choose a loss function $\mathcal{L}$ that mathematically specifies our objective. We then optimize $\alpha$ to match the target; specifically, we iteratively 1) simulate the trajectory of the soft structure given the current $\alpha$, 2) use DiffPD to compute model-based gradients $\frac{\partial \mathcal{L}}{\partial \alpha}$, and 3) perform an optimization step to update the guess for $\alpha$ using L-BFGS.  We repeat this process until convergence.}

\section{Results}\label{results}
\subsection{Clamped \rkkf{Soft} Beam under External Force}\label{results_beam}

Studying a clamped beam structure allows us to solve for certain variables and thus reduce the number of factors that influence more complex simulations. 
For this mechanical structure, we focus on the vertical deformation of the beam's tip ($z$ axis
). Here, we present the results for the point ``Left'' as defined in \Cref{fig:beam_scheme}. The results for point ``Right'' are similar and therefore not shown here.

\subsubsection{Meshing, Position Tracking, and Steady State Comparison}

We experiment with \emph{three} loading scenarios in this section (\Cref{tab:beam_load_scenarios}). These scenarios were obtained by varying the edge force $F$
. Additionally, we distinguish cases by their number of Degrees of Freedom (DOFs) and by the type of mesh elements that were used; namely, hexahedra (``Hex Mesh'') or tethrahedra (``Tet Mesh'').

\begin{table}[htb]
\centering
\begin{tabular}{@{}lllll@{}}
\toprule
Case & Load Scenario & DOFs Hex & DOFs Tet & DOFs COMSOL\\ \midrule
A-1  & $F=0$ N            & 4608         & 3834         & 7110       \\
A-2  & $F=0$ N            & 4608         & 183516       & 7110       \\
B    & $F=0.510$ N      & 4608         & 3834         & 7110       \\
C    & $F=0.510$ N      & 9900         & 71688        & 7110       \\
D    & $F=0.991$ N      & 4608         & 3834         & 7110       \\
E    & $F=0.991$ N      & 9900         & 71688        & 7110       \\ \bottomrule
\end{tabular}
\caption{Summary of the loading cases for the clamped beam.}
\label{tab:beam_load_scenarios}
\end{table}

\begin{figure}[tb]
    \vspace{0.5em}
    \centering
    \includegraphics[width=0.96\linewidth]{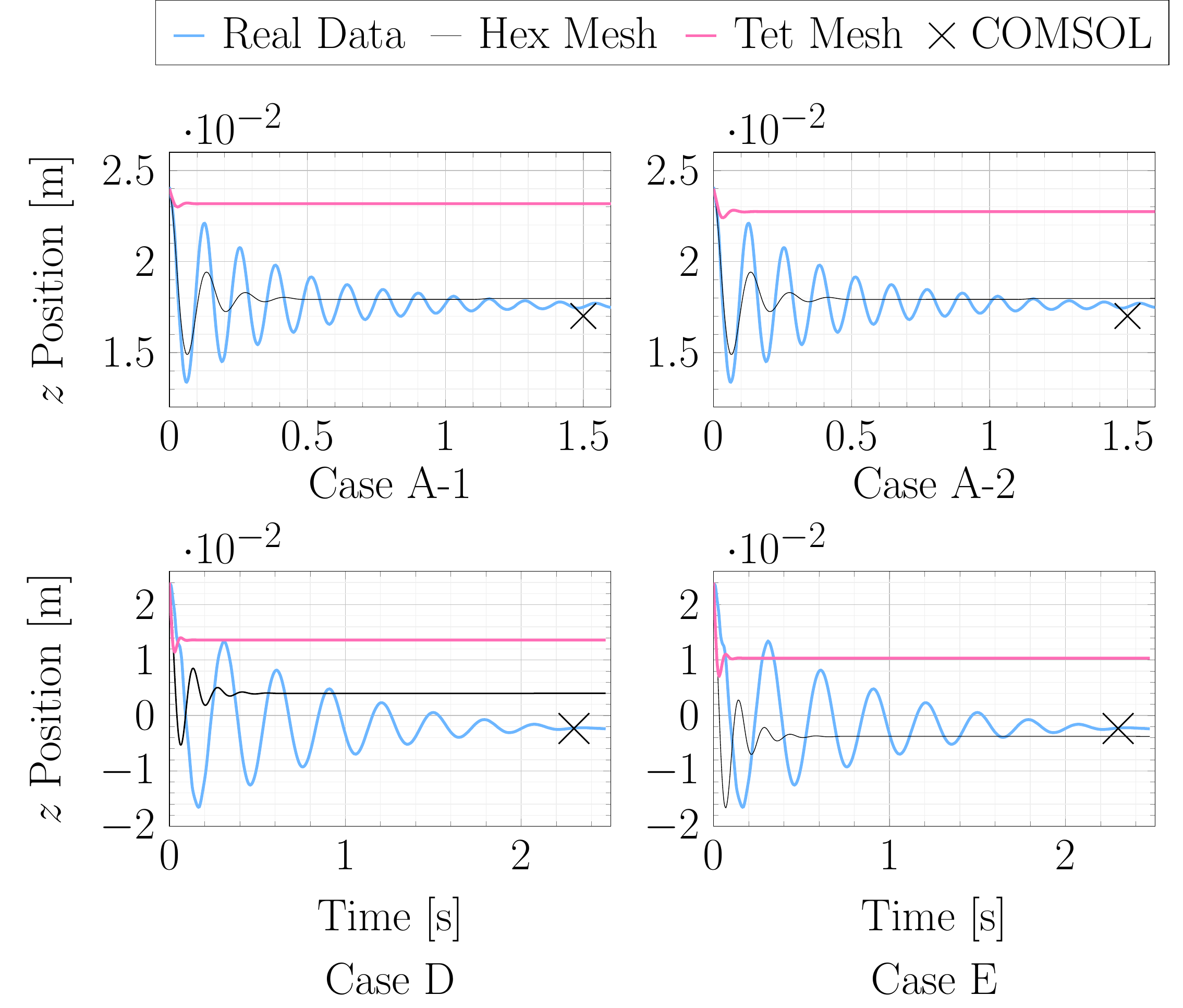}
    \caption{Comparison of real data and simulated results using the Hex Mesh and the Tet Mesh, along with the COMSOL static solution.}
    \label{fig:beam_results_steady_state}
\end{figure}

For Case A-1, we find close correspondence between the real data, the COMSOL solution, and the Hex Mesh solution: all three steady state solutions are within $1$ \si{mm} of each other. On the other hand, the Tet Mesh response is too stiff; it has $6 $ \si{mm} \md{absolute error} 
(Fig. \ref{fig:beam_results_steady_state} Case A-1). 

In Case A-2, we verify if the Tet Mesh behavior is an artifact stemming from a poor choice of the mesh resolution. We perform the same simulation again while increasing the number of DOFs for the Tet Mesh from 3834 to 183516. As shown in Fig. \ref{fig:beam_results_steady_state} Case A-2, only a very small difference ($0.438$ \si{mm} difference) can be observed.

Keeping the same number of DOFs for the two mesh types as in Case A-1, we add an edge force at the tip of the beam for Case B and D. Under this configuration, the simulation yields incorrect results for both mesh types. However, increasing the number of DOFs of the Hex Mesh resolves this issue and leads to more accurate results. For the Tet Mesh, this refining step is not enough to elicit a sufficient response to match real data. \md{We only present the results for Case D and E in Fig. \ref{fig:beam_results_steady_state}, as Case B and C exhibit similar results.} 
\md{Since the DiffPD simulation for case E with the Hex Mesh is accurate, we can use the differentiability of the framework to calibrate the Young's Modulus. Following the steps described in \Cref{optimization_algorithm} with 
$\mathcal{L}=\| \bm{q}_{\text{sim,final}} - \bm{q}_{\text{real,final}} \|^2$ and for different initial values, the average optimized Young's Modulus is $E=283271$ \si{Pa}, which is close to the solution we found using COMSOL ($E=263824$ \si{Pa}, \Cref{simulation_setup}).}


\subsubsection{Numerical Damping by the Implicit Euler Method}\label{beam_numerical_damping}

\paragraph{Analytical Solution}\label{numerical_damping_analytical_solution}
As introduced in \Cref{material_damping}, we can describe the damping of an oscillation using the damping ratio $\zeta$. Here, we show an analytical relationship between the implicit Euler time step $h$ and the induced numerical damping, characterized by the damping ratio $\zeta$.

We consider the system \eqref{numerical_damping_example}, whose exact solution is the free undamped vibration \eqref{numerical damping_example_exact_solution}. We can reformulate the solution obtained using the implicit Euler method \eqref{euler_induction_formula} as:
\begin{equation}
     x_n= \left(\frac{1}{1-hi\omega}\right)^n = \frac{(1+\omega h i)^n}{(1+\omega^2 h^2)^n}=\frac{e^{n\cdot \angle(1+\omega hi)}}{(1+\omega^2 h^2)^{n/2}}
\end{equation}

The values of the positive peaks of this oscillation can be found at step $n^\ast$ as follows:
\begin{equation}
    n^\ast \cdot \angle(1+\omega hi) = k \cdot 2\pi\; \Rightarrow \;  n^\ast = \frac{k \cdot 2\pi}{\angle(1+\omega hi)},\; k\in\mathbb{N}
\end{equation}

The amplitude of the oscillation at $n^\ast$ is
\begin{equation}
    X_k = \frac{1}{(1+\omega^2 h^2)^{n^\ast/2}} =  \frac{1}{(1+\omega^2 h^2)^{\frac{k\pi}{\angle(1+\omega hi)}}}
\end{equation}

To express the damping ratio $\zeta$, we first need to find an expression for the logarithmic decrement $\delta$ \aes{\eqref{log_dec}}:
\begin{equation}
    \begin{aligned}
        \delta(h) &= \frac{1}{m}\cdot \ln\left(\frac{X_k}{X_{k+m}}\right)\\
               &= \frac{1}{m}\cdot \ln\left\{\frac{(1+\omega^2 h^2)^{\frac{(k+m)\pi}{\angle(1+\omega hi)}}}{(1+\omega^2 h^2)^{\frac{k\pi}{\angle(1+\omega hi)}}}\right\}\\
               &= \frac{\pi}{\angle(1+\omega hi)}\cdot \ln \left(1+\omega^2 h^2\right)
    \end{aligned}
\end{equation}

Knowing the logarithmic decrement $\delta$, we can write the damping ratio $\zeta$ \aes{as a function of $h$} using \eqref{damping_ratio}:
\begin{equation}\label{damping_ratio_analytical_solution_equation}
    \zeta(h)=\frac{\delta(h)}{\left(4\pi^2+\delta(h)^2\right)^{1/2}}
\end{equation}

\begin{figure}[tb]
    \vspace{0.5em}
    \centering
    \includegraphics[width=0.95\linewidth]{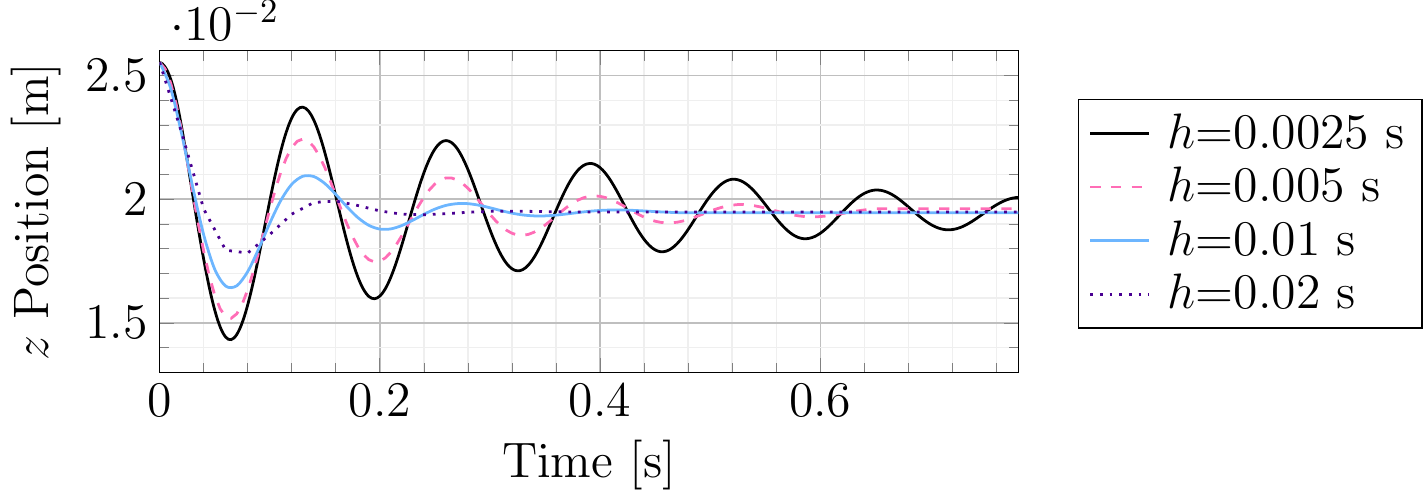}
    \caption{Simulation results of Case A-1 for a selection of different time steps $h$.}
    \label{fig:numerical_damping_response_gravity}
\end{figure}

\paragraph{Numerical Damping in \mdf{FEM}}\label{beam_numerical_damping_diffpd}
To experimentally describe the relation between $\zeta$ and  $h$, we simulated the Hex Mesh beam for different $h$ (Fig. \ref{fig:numerical_damping_response_gravity}). 

\begin{figure}[tb]
    \centering
    \includegraphics[width=0.96\linewidth]{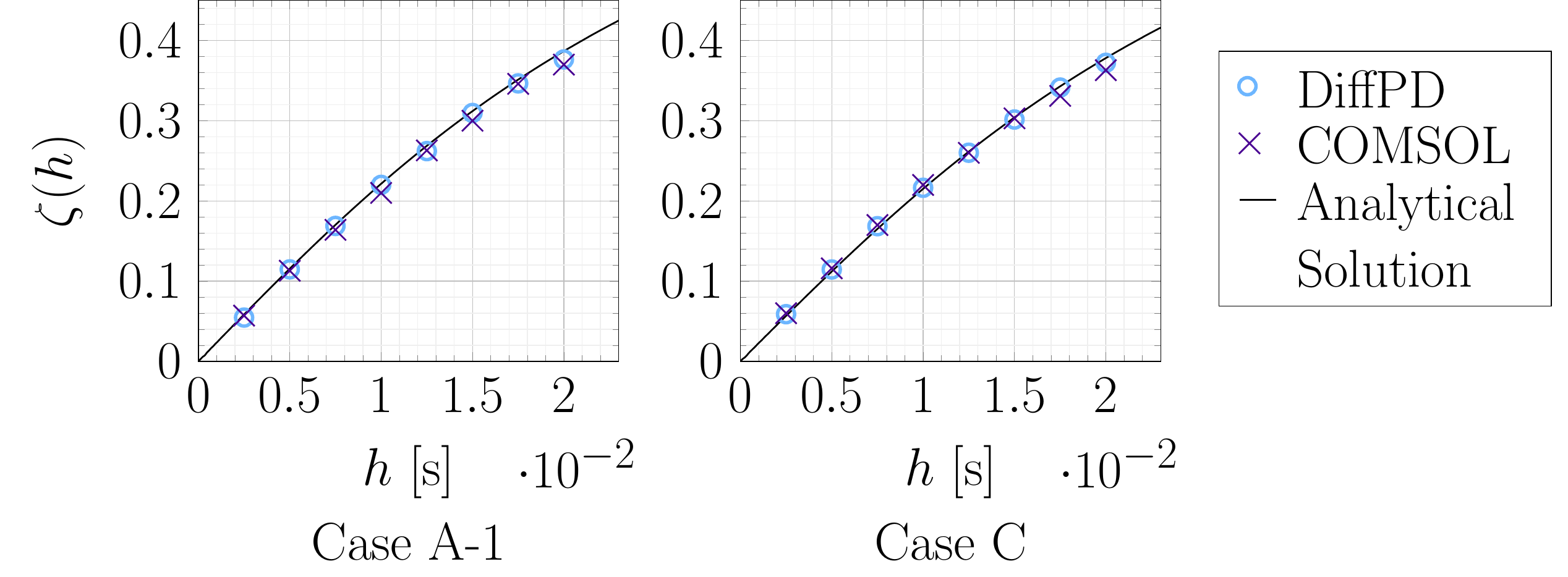}
    \caption{Comparison between the computed values of $\zeta(h)$ from the DiffPD and COMSOL (using implicit Euler) simulations and our analytical solution.}
    \label{fig:numerical_damping_diffpd_vs_analytical}
\end{figure}

With the values we gathered from the simulations, we can determine $\zeta$ \md{using \eqref{log_dec} and \eqref{damping_ratio}}. The vertical displacement of the tip points of the beam can be approximated as a single DOF vibration, and we can observe that our analytical solution \eqref{damping_ratio_analytical_solution_equation} closely matches the value computed from the measurement data (Fig. \ref{fig:numerical_damping_diffpd_vs_analytical}). \md{We note that this result generalizes to other FEM simulators using implicit Euler solvers, as shown by simulation results using COMSOL with implicit Euler method.}

\subsubsection{Material Damping}\label{beam_material_damping}

\begin{figure}[tb]
    
    \centering
    \includegraphics[width=0.9\linewidth]{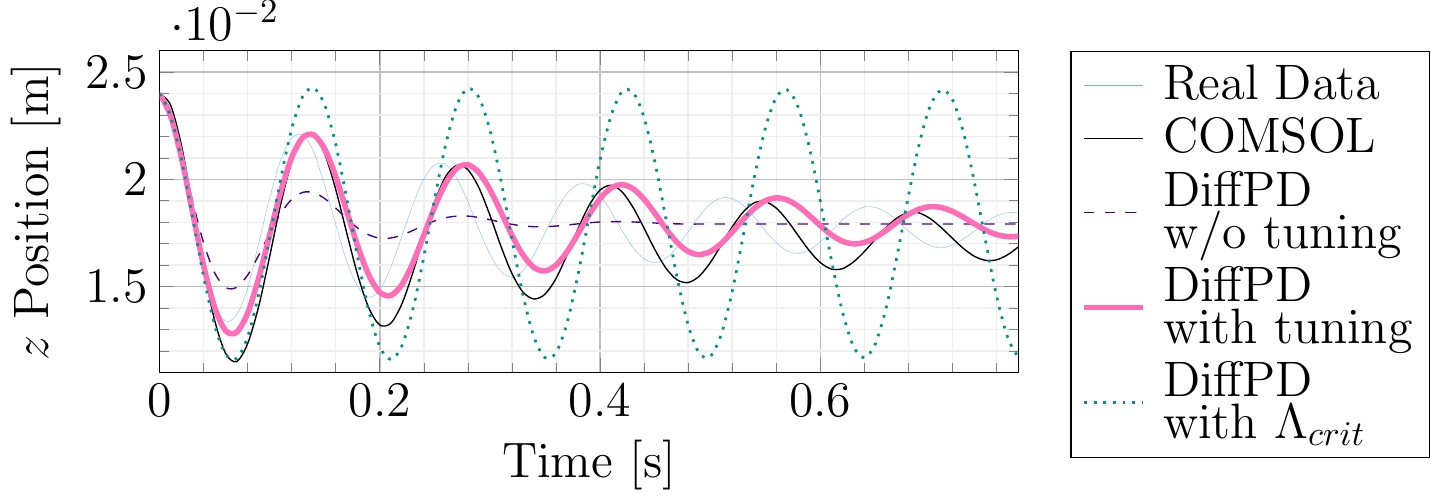}
    \caption{Material damping modeling results using a velocity dependent state force.}
    \label{fig:Damping_compensation}
\end{figure}

\begin{figure}[tb]
    
    \centering
    \includegraphics[width=1\linewidth]{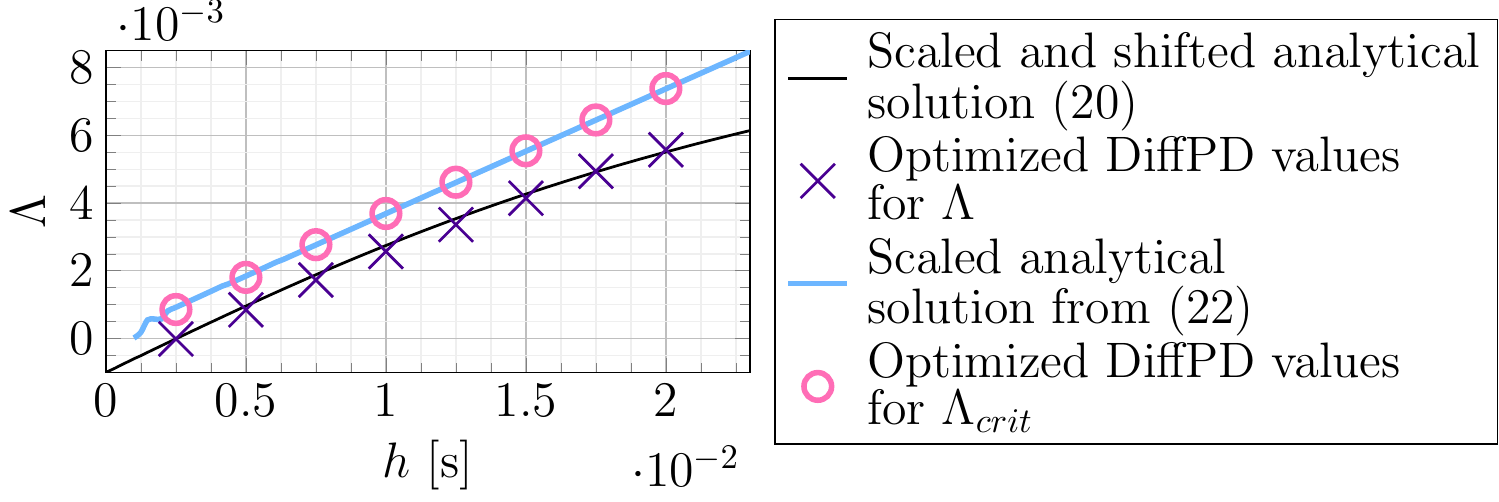}
    \caption{\mdf{$\Lambda$ and $\Lambda_{{crit}}$ as a function of $h$.}}
    \label{fig:Damping_compensation_lambda}
\end{figure}

\paragraph{\mdf{Matching the real-world dynamic response}}\label{beam_material_damping_matching}
As introduced in \Cref{modeling_damping}, we let the numerical damping be the only source of damping in the \mdf{FEM} simulation \md{and use the differentiability of DiffPD to find} $\Lambda$ so that the \aes{simulated} solution closely approximates the real-world physical response. \mm{We define the optimization objective to be that the simulated dynamic oscillation should match the envelope of the real oscillation (Fig. \ref{fig:Damping_compensation})}.
Plotting  the tuning parameter $\Lambda$ as a function of time step $h$, a similar relation to the one between $\zeta$ and $h$ \eqref{damping_ratio_analytical_solution_equation}, can be observed (Fig. \ref{fig:Damping_compensation_lambda}).

\paragraph{\mdf{Stability Analysis}}\label{beam_material_damping_stability} \mdf{The notion of A-stability describing the implicit Euler method is not sufficient to assess the system's stability when active forces are considered. The velocity-dependent force \eqref{damping_compensation_formula} can lead the simulations to become unstable if $\Lambda$ is larger than $\Lambda_{{crit}}$. To illustrate this, we derive an analytical criterion based on the linear system \eqref{numerical_damping_example} that includes a velocity-dependent force:}

\mdf{
\begin{equation}\label{CT_equation_damping}
    \frac{d}{dt}\begin{bmatrix}x\\\dot{x}\end{bmatrix}=
    \begin{bmatrix}0& 1\\-\omega^2& \Lambda\end{bmatrix}\begin{bmatrix}x\\\dot{x}\end{bmatrix}
\end{equation}}

\mdf{Using the implicit Euler method on \eqref{CT_equation_damping} leads to a discrete system of the form $\underline{x}_{j+1}=A \underline{x}_{j}$:}

\mdf{
\begin{equation}\label{difference_equation_damping_linear_form}
     \begin{bmatrix}x_{j+1}\\v_{j+1}\end{bmatrix}=\frac{1}{1-h\Lambda+h^2\omega^2}\begin{bmatrix}1-h\Lambda &h \\
     -h\omega^2 & 1\end{bmatrix}   \begin{bmatrix}x_{j}\\v_{j}\end{bmatrix}
\end{equation}}

\mdf{\eqref{difference_equation_damping_linear_form} is stable iff the eigenvalues of $A$ are in the unit circle. This is the case if $\Lambda<\Lambda_{{crit}}$. $\Lambda_{{crit}}$ can be represented as a function of $h$ (Fig. \ref{fig:Damping_compensation_lambda}). The corresponding values from the DiffPD simulations can be obtained \mmf{using their differentiable property} by defining an optimization objective that requires that the oscillation has constant amplitude. The obtained values closely follow our analytical solution up to a constant factor, which is likely primarily due to the force being applied at multiple nodes, and not at only one point particle as in \eqref{difference_equation_damping_linear_form}.}

\subsection{Soft Robotic Arm}\label{results_arm}\label{actuation_mucles_models}
In this section, we simulate a more complex structure that includes pneumatic actuation, namely, a soft robotic arm.

\begin{figure}[tb]
    \centering
    \includegraphics[width=0.96\linewidth]{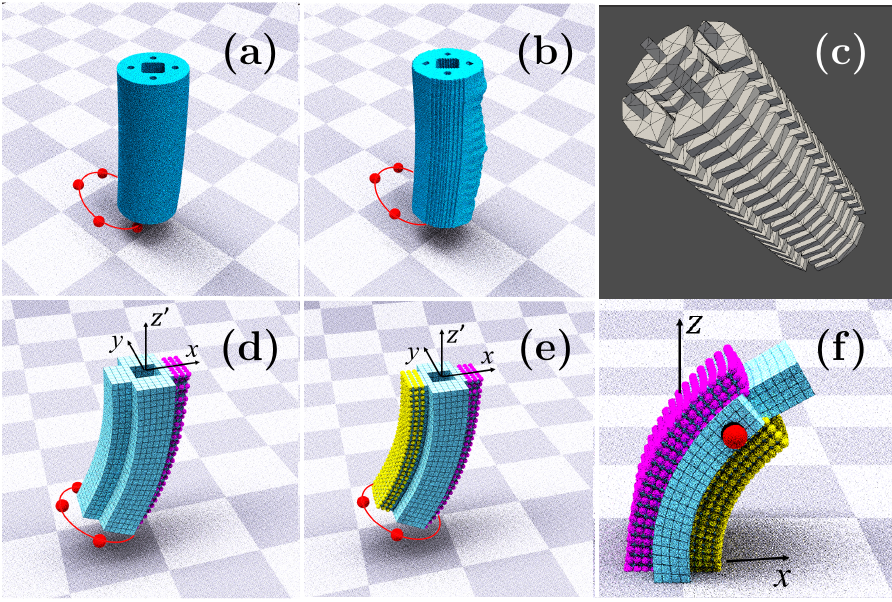}
    \caption{Simulation results for the soft robotic arm and soft fish tail. The red dots are the motion markers positions of the real robot. The pink elements correspond to muscle fibers in extension, and the yellow elements to muscle fibers in contraction.}
    \label{fig:arm_simulation_results}%
\end{figure}

The complex inner geometry of the pressure chambers is difficult to model accurately. \aes{Their modeling requires a very fine mesh (Fig. \ref{fig:arm_simulation_results} (c)) and is therefore computationally costly}. A Tet Mesh suffers from the same stiff behavior as the beam simulations in \Cref{results_beam} (Fig. \ref{fig:arm_simulation_results} (a)).
Meanwhile, a Hex Mesh cannot precisely capture the fine ribbing geometry (Fig. \ref{fig:arm_simulation_results} (b)) \aes{due to the regular shape of its elements}. 

\begin{figure}[tb]
    \centering
    \includegraphics[width=0.98\linewidth]{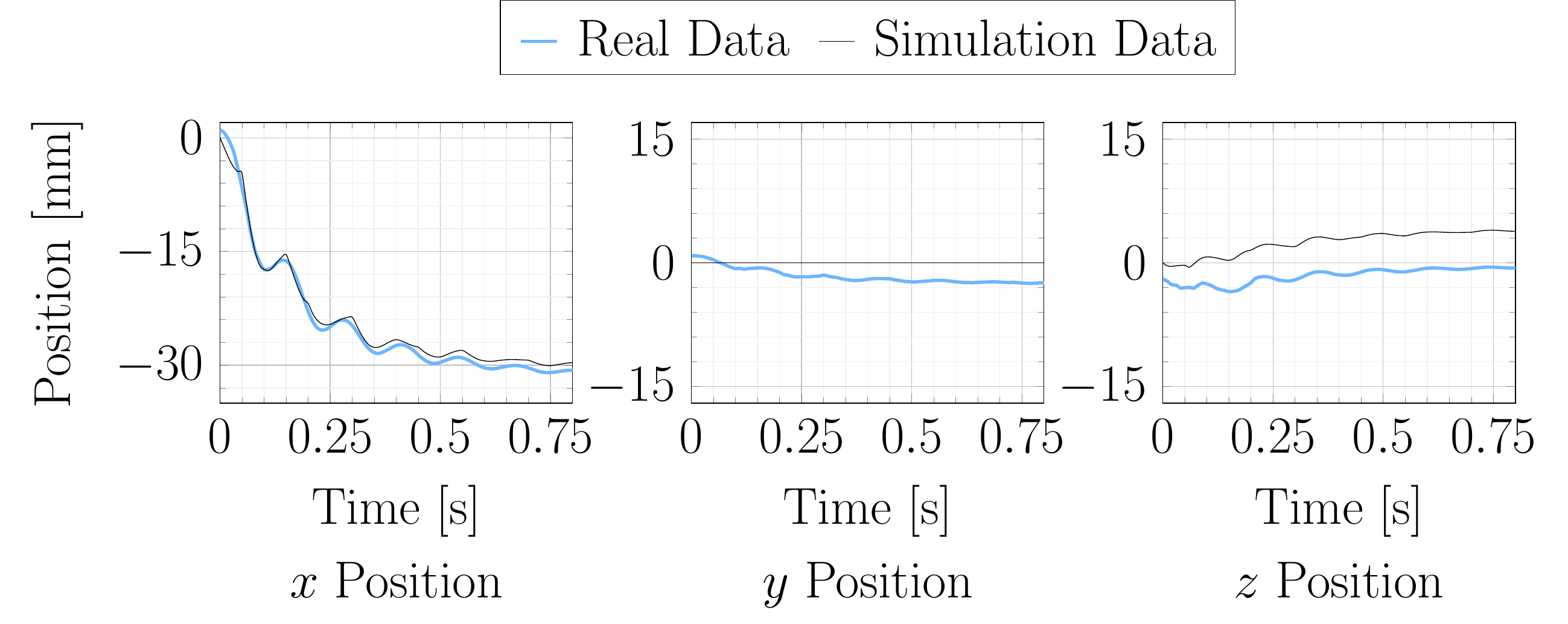}
    \caption{Detailed simulation results for the Muscle Model AC2. We optimize for the $x$ position. The difference we observe for the $y$ and $z$ position probably stems from the small manufacture inaccuracies of the arm or experimental setup.}
    \label{fig:AC2_results}%
    \vspace{-5pt}
\end{figure}

As an \md{alternative} solution, we propose 
lower-order ``Muscle Models'' to simulate the arm. These models are based on the ``Muscle Energy'' model, introduced in \Cref{modeling}. 
We specify ``muscle fibers'' along a simplified hexahedral mesh  (Fig. \ref{fig:arm_simulation_results}), along which every hexahedron in the fiber is given the same actuation signal, as a means of emulating pneumatic behavior. We compare the performance of two possible muscle configurations in Fig. \ref{fig:AC2_results}: (d) AC1 has a single muscle, which extends, while (e) AC2 has an extending muscle and an antagonistic contracting muscle. \aes{Muscle Models use considerably less memory and are faster to simulate than meshing the true ribbing of the actuator. Furthermore, they do not need to resolve complex internal self-collisions.}

We consider two experiments for each arm design, each of which leverages the differentiable character of DiffPD to optimize the muscle actuation signals $a$.  In the first experiment, we optimize a single actuation signal, applied at every time step, 
such that the final pose of the virtual arm matches a static physical arm pneumatically actuated with a fixed constant pressure. In the second experiment, we optimize an actuation sequence to match \md{our simulations} with the full trajectory of the physical arm
. This actuation sequence applies actuation signals at a fixed frequency $f$;
if our simulation runs for $T_{total}$ seconds, the actuation sequence has $\lfloor \frac{T_{total}}{f} \rfloor$ decision variables. 

\begin{figure}[tb]
    \vspace{0.5em}
    \centering
    \includegraphics[width=0.96\linewidth]{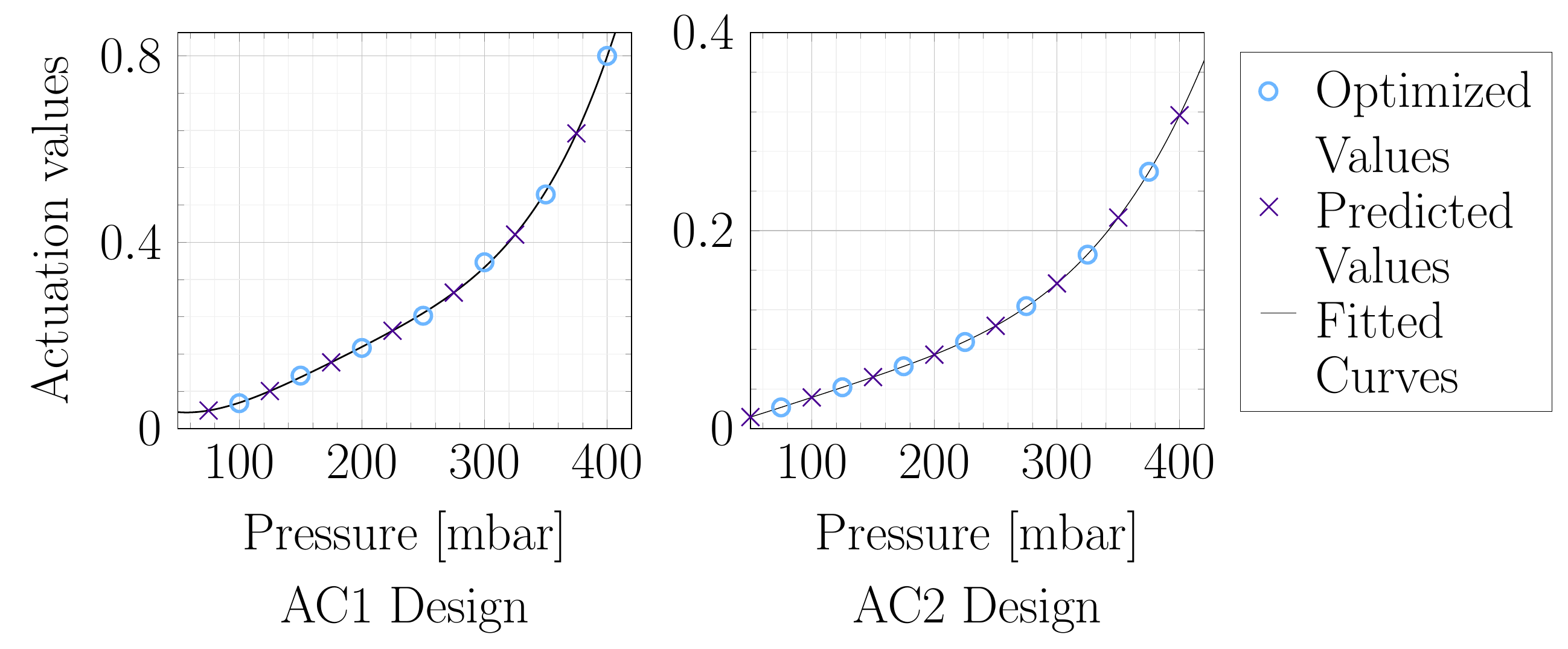}
    \caption{Pressure-to-actuation maps for the AC1 and AC2 designs. 
    }
    \label{fig:pressure_actuation_map}
\end{figure}

In the single-actuation , we optimize the loss-function $\mathcal{L} = \frac{1}{T \cdot M} \sum_{t=1}^T \sum_{i=1}^M \| \bm{q}_{\text{sim}, i}(t) - \bm{q}_{\text{real}, i}(T) \|^2$ where $\bm{q}_i$ is the position of motion marker $i$; this prioritizes that the soft arm achieves an accurate static response. Note that $\mathcal{L}$ is dependent on $a$ since the final pose $\bm{q}_{\text{sim}}$ depends on $a$. By repeating optimization for successive pressure values, we extract pressure-to-actuation maps (Fig. \ref{fig:pressure_actuation_map}). \md{By fitting a curve through the optimized actuation values, we can predict interpolated values for which we did not optimize.  For these predicted actuation values there is high agreement between the physical arm and simulated models, with an average relative position error of $13.07 \%$}. 

In the multi-actuation example, we optimize the similar loss $\mathcal{L} = \frac{1}{T \cdot M} \sum_{t=1}^T \sum_{i=1}^M \| \bm{q}_{\text{sim}, i}(t) - \bm{q}_{\text{real}, i}(t) \|^2$ (note the term $\bm{q}_{\text{real}, i}(t)$ cf $\bm{q}_{\text{real}, i}(T)$); this promotes dynamic position tracking over time. In our experiments, the actuation frequency is 20Hz. Fig. \ref{fig:AC2_results} shows that the optimized actuation sequence accurately matches the tracked physical arm.

These results demonstrate that Muscle Models can be accurately and predictively mapped to real pressure actuation.  

\subsection{Pneumatically Actuated Fish Tail}\label{results_fish}
\begin{figure}[tb]
    \centering
    \includegraphics[width=0.98\linewidth]{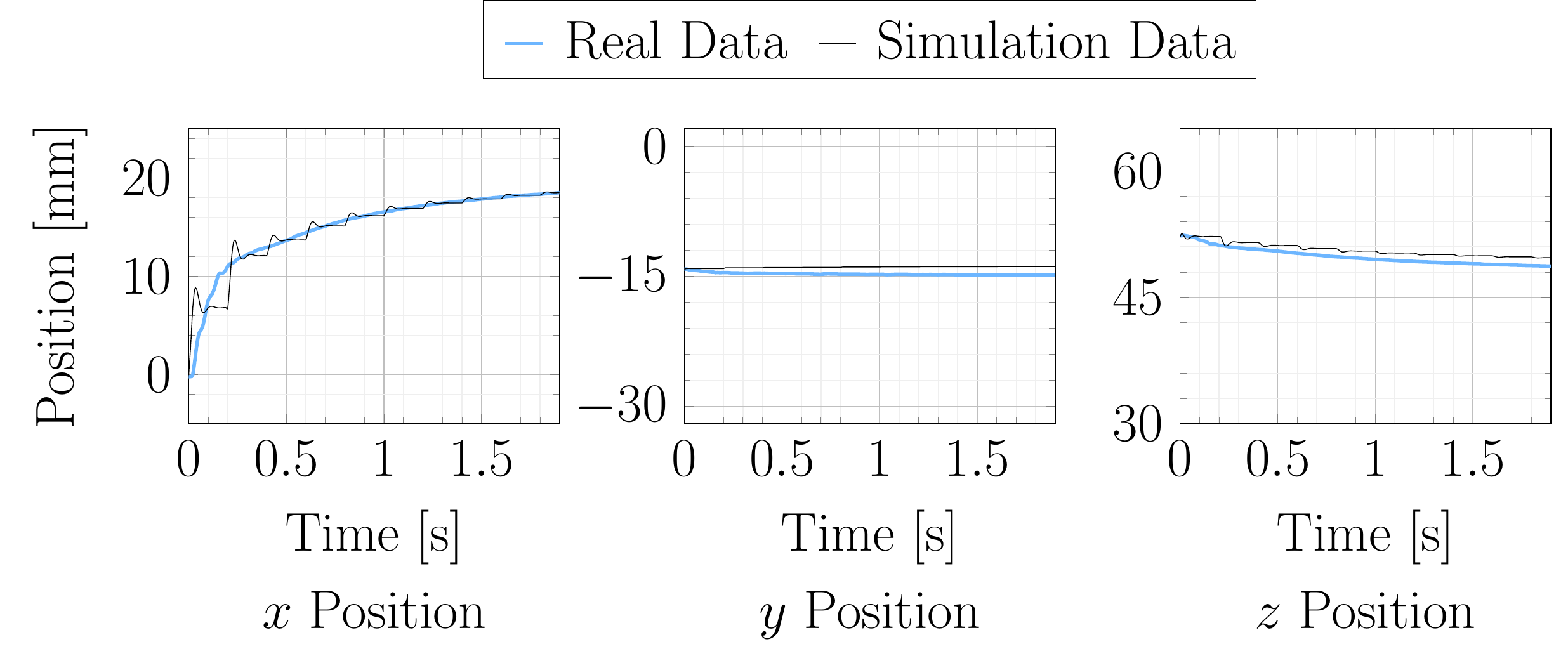}
    \caption{Detailed simulation results for the Muscle Model of the fish. We optimize trajectories to match target $x$ and $z$ positions.}
    \label{fig:fish_results}%
\end{figure}

To validate our Muscle Models on a separate geometry, we model the pneumatically actuated soft robotic fish tail (\Cref{experimental_setup}). As with the arm, our Muscle Models' actuations are optimized to follow the dynamic behavior of the real-world robot (Fig.~\ref{fig:fish_results}). We optimized for actuation sequences with 5Hz control signals. 
Despite the step-wise behavior of the simulation, 
our optimized solution is still able to very closely match the motions of this second robotic morphology.

\section{Discussion and Future Work}\label{discussion_and_conclusion}


In this paper, we investigated several traits of \mdf{FEM} modeling
, and provided recommendations on how to better 
match physical reality and overcome the sim-to-real gap.

Our first recommendation resulting from the experiments is with regards to meshing.
The linear tetrahedral mesh elements produced inaccurate behavior in the incompressible regime. This problem, named \emph{locking problem}, is well known in the FEM community. 
On the other hand, the steady state results of the DiffPD simulations using the Hex Mesh are satisfying \md{and allow for parameter calibration.} 
Therefore, we recommend \rkkf{to use \mdf{a Hex Mesh} for problems in the incompressible regime.} 

Our second recommendation is with respect to damping.  Although the steady state results are simple to match precisely once system identification of the Young's modulus is achieved, the dynamic behavior of vanilla \mdf{FEM} simulation accumulates significant error. \mdf{DiffPD}, using the implicit Euler method, suffers \rkkf{from} numerical damping \mdf{as derived analatytically. } 
Depending on the exact use of \mdf{FEM} simulations, it is crucial to be aware of numerical damping, and a linear damping model is recommended to fit the material's response.

Our third recommendation is with respect to actuator modeling.  Complex structures, such as the ribbed pressure chambers \rkkf{of fluidic actuators}, require a fine mesh to be modeled correctly. These complex geometries can generally not be modeled with a Hex Mesh, due to the regular shape of its elements. Although a \md{linear} Tet Mesh is able to represent the geometry of the pressure chambers of the arm correctly, it fails to model the deformation of the arm correctly \md{and is computationally costly}. 
Muscle Models, correctly designed, can lead to good results in accuracy and computational efficiency. The differentiability of DiffPD allows efficient optimization of the actuator behavior.


\rkkf{We see four areas of }further investigation.  
First, to model \rkkf{a} more complex geometry without experiencing element locking, higher-order tetrahedral elements should be considered.  Second, since the Muscle Models show promise as a means of modeling \rkkf{fluidic} actuation, it would be useful to develop more general, data-driven rules to obtain pressure-to-actuation maps for other geometries. 
\aesf{Third, while our methodology and its principles are general}\rkkf{ly applicable} \aesf{to any finite-element-based simulator, DiffPD was primarily employed in this work}.   
\aesf{Further validation with other FEM simulators would be valuable.}
Finally, investigating 
more nonlinear materials or dynamics, in which Neo-Hookean responses play a role, would be helpful for future applications.

\addtolength{\textheight}{0cm}   







\bibliographystyle{IEEEtran}
\bibliography{IEEEabrv,references.bib, sim2realbib.bib}

\end{document}